\documentclass[pdflatex,sn-mathphys-num]{sn-jnl}

\usepackage{graphicx}%
\usepackage{multirow}%
\usepackage{amsmath,amssymb,amsfonts}%
\usepackage{amsthm}%
\usepackage{mathrsfs}%
\usepackage[title]{appendix}%
\usepackage{xcolor}%
\usepackage{textcomp}%
\usepackage{manyfoot}%
\usepackage{booktabs}%
\usepackage{algorithm}%
\usepackage{algorithmicx}%
\usepackage{algpseudocode}%
\usepackage{listings}%

\usepackage{comment}
\usepackage{wrapfig}
\usepackage{subfigure}
\usepackage{adjustbox}

\usepackage{soul, color}
\usepackage{colortbl}
\definecolor{Gray}{gray}{0.9}

\newcommand{\ve}[1]{\mathbf{#1}} 
\newcommand{\tve}[1]{\tilde{\mathbf{#1}}} 
\newcommand{\bve}[1]{\bar{\mathbf{#1}}} 

\DeclareMathOperator*{\argmin}{arg\,min}
\DeclareMathOperator{\arctantwo}{arctan2}

\begin{document}

\title[ORACLE-Grasp: Zero-Shot Affordance-Aligned Robotic Grasping using Large Multimodal Models]{ORACLE-Grasp: Zero-Shot Affordance-Aligned Robotic Grasping using Large Multimodal Models}

\author[1]{\fnm{Avihai} \sur{Giuili}}\email{avigiuili@mail.tau.ac.il}

\author[1]{\fnm{Rotem} \sur{Atari}}\email{rotematari@mail.tau.ac.il}

\author*[1]{\fnm{Avishai} \sur{Sintov}}\email{sintov1@tauex.tau.ac.il}

\affil*[1]{\orgdiv{School of Mechanical Engineering}, \orgname{Tel-Aviv University}, \orgaddress{\street{55 Chaim Lebanon St.}, \city{Tel-Aviv}, \postcode{6997801}, \country{Israel}}}


\abstract{Grasping unknown objects in unstructured environments is a critical challenge for service robots, which must operate in dynamic, real-world settings such as homes, hospitals, and warehouses. Success in these environments requires both semantic understanding and spatial reasoning. Traditional methods often rely on dense training datasets or detailed geometric modeling, which demand extensive data collection and do not generalize well to novel objects or affordances. We present ORACLE-Grasp, a zero-shot framework that leverages Large Multimodal Models (LMMs) as semantic oracles to guide affordance-aligned grasp selection, without requiring task-specific training or manual input. The system reformulates grasp prediction as a structured, iterative decision process, using a dual-prompt tool-calling strategy: the first prompt extracts high-level object semantics, while the second identifies graspable regions aligned with the object’s function. To address the spatial limitations of LMMs, ORACLE-Grasp discretizes the image into candidate regions and reasons over them to produce human-like and context-sensitive grasp suggestions. A depth-based refinement step improves grasp reliability when available, and an early stopping mechanism enhances computational efficiency. We evaluate ORACLE-Grasp on a diverse set of RGB and RGB-D images featuring both everyday and AI-generated objects. The results show that our method produces physically feasible and semantically appropriate grasps that align closely with human annotations, achieving high success rates in real-world pick-up tasks. Our findings highlight the potential of LMMs for enabling flexible and generalizable grasping strategies in autonomous service robots, eliminating the need for object-specific models or extensive training.}

\keywords{Grasping, Large Multimodal Models, Affordance-aligned Grasping}



\maketitle

\section{Introduction}
\label{sec:introduction}

Robotic grasping remains a core challenge in autonomous manipulation, particularly for service robots operating in unstructured and dynamic environments such as homes, hospitals or warehouses. Traditional approaches often depend on precise object models, depth-based geometry recognition, or supervised learning pipelines trained on large-scale datasets \cite{Eppner2015, Levine2018}. However, real-world deployment frequently requires interacting with novel objects that exhibit unpredictable shapes and require task-based reasoning. In such scenarios, accurate grasp localization, identifying grasp points and orientations that are both physically feasible and task-appropriate, becomes critical \cite{Zeng2018}. While recent deep learning methods have advanced grasp quality prediction \cite{Mahler2019}, they primarily focus on geometric properties and neglect semantic reasoning about the object or intended task. This gap motivates the development of new frameworks that combine high-level semantic understanding with spatial reasoning, enabling more human-like, adaptive grasp selection in the real-world.

\begin{figure}
    \centering
    \includegraphics[width=\linewidth]{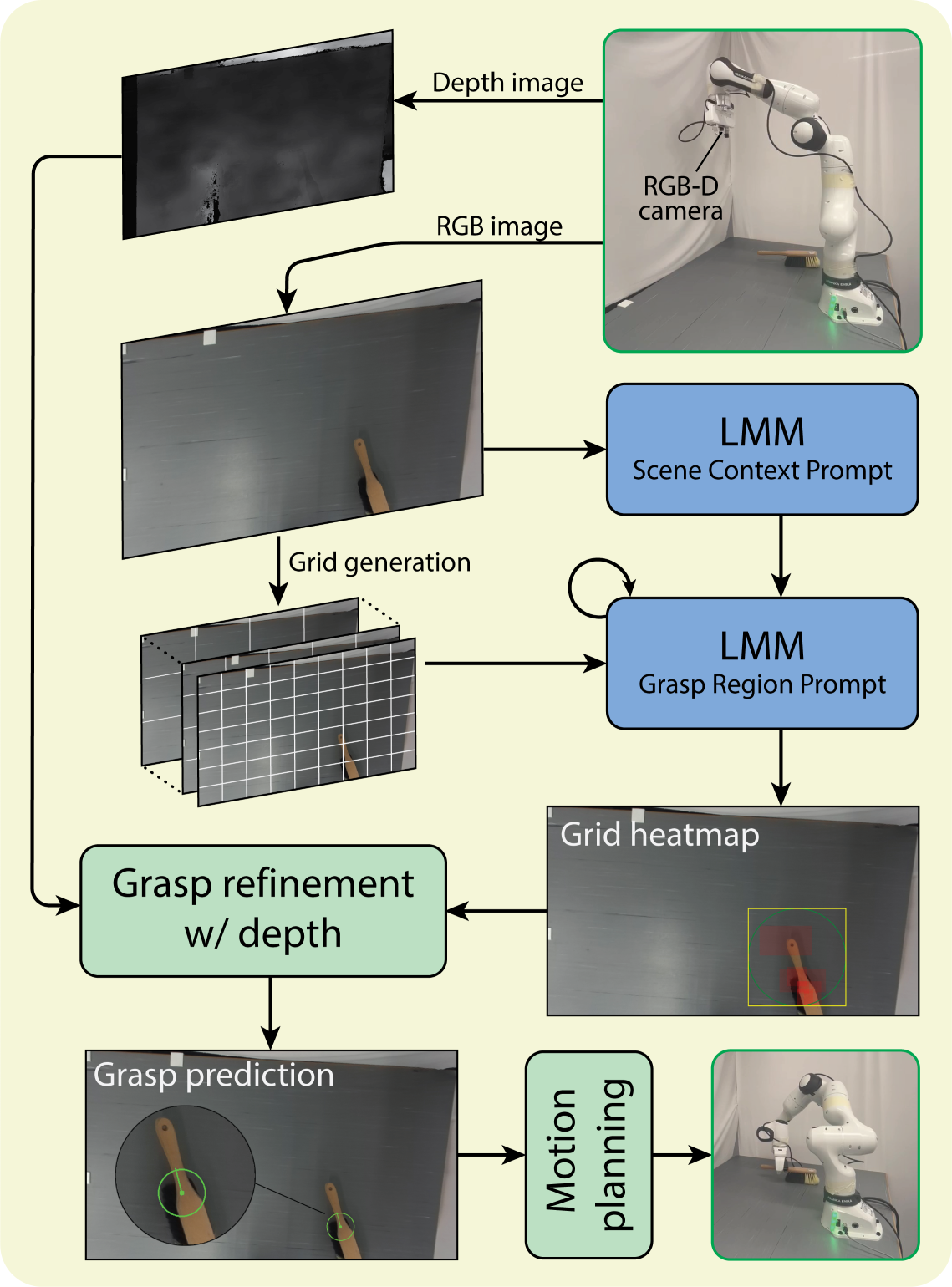}
    \caption{Overview of the ORACLE-Grasp framework. An RGB-D camera captures the scene. An LMM uses dual prompts over a grid to identify a graspable region, which is then refined with depth data for precise grasp prediction and executed by a robot.}
    \label{fig:intro}
    \vspace{-0.5cm}
\end{figure}

Previous approaches have typically relied on geometric analysis of depth data \cite{Bohg2014,Huang2024}, model-based grasp synthesis \cite{Sahbani2012} or extensive training of deep networks on large-scale grasp datasets \cite{Morrison2018,Khor2024}. Early work on learning-based methods for grasping novel objects used machine learning techniques to predict effective grasp points directly from 2D images without requiring complete 3D reconstruction or object recognition \cite{Saxena2008}. However, deep learning-based approaches typically focus on learning and predicting the quality of candidate grasps, without integrating semantic reasoning about the task or the object’s context \cite{Mahler2017}. While recent methods have made significant advances in pixel-wise grasp prediction and affordance learning \cite{Shao2020, Zeng2022}, they often lack the semantic reasoning needed for affordance-aligned grasping, especially in zero-shot settings where the object has not been seen before. Grasp prediction in real-world contexts often involves more than simply securing a stable hold; it requires selecting grasp points that are appropriate for the intended use of the object. For example, while a hammer can be grasped by either the head or the handle from a geometric perspective, semantically meaningful use (e.g., hammering a nail or handing it to a human) strongly favors the handle. Semantic reasoning helps resolve such ambiguities by considering object function and affordance, which are crucial for selecting grasps that are not only physically robust but also task-relevant and human-like \cite{Dang2012}. Without this layer of understanding, grasping methods risk producing functionally incorrect or unsafe outcomes.

Large Language Models (LLMs) \cite{Minaee2024} and Large Multimodal Models (LMMs) \cite{wang2024} have revolutionized robotics by enabling advanced natural language understanding and multimodal perception \cite{wang2024large}. These powerful models empower robots to interpret complex instructions, reason about their environment, and perform tasks with greater autonomy and flexibility. Despite notable advances of LMMs in high-level robotic task planning and embodied reasoning, their application to robotic grasping remains less explored. Most existing work leveraging LMMs in robotics focuses on semantic instruction following, policy learning or general manipulation planning, rather than on fine-grained grasp localization and selection.

Existing approaches include PhyGrasp \cite{guo2024}, which introduces an LMM that integrates 3D point clouds and language to inform grasping decisions based on estimated physical properties such as material, fragility and mass. However, PhyGrasp requires detailed 3D shape modeling and the availability of physical property annotations. It also depends on large-scale data collection and does not explicitly address task-driven, human-like grasp selection from single-view visual inputs. Another emerging line of work employs LMMs to direct grasping based on human verbal instructions, providing external context and object knowledge to guide the robot’s actions \cite{Tziafas2023,Jin2024, Xu2024, li2024, Li2024ShapeGrasp}. While these approaches support task-specific grasping, they depend on human guidance at inference time and often require pretraining on large, specialized grasp datasets, limiting their scalability for fully autonomous deployment. Moreover, they focus on selecting which object to grasp, rather than identifying affordance-aligned, human-like grasp locations on the object itself. Furthermore, these methods often rely on supervised components for grasp detection and quality evaluation, lack explicit spatial reasoning, and exhibit limited generalizability as they are typically trained on a fixed set of object classes, thereby hindering truly zero-shot execution \cite{Tang2023b, Tang2025}.

In this work, we investigate the ability of LMMs to generate context-free affordance-aligned grasps, i.e., grasps that do not rely on explicit object or task grounding provided by the user, but instead reflect the most probable or conventional way an object would be grasped to perform its likely function. For example, a knife or drill is typically grasped by the handle, regardless of explicit instruction. Unlike prior approaches that require structured input or custom-trained grasp encoders \cite{minderer2022simple, Tang2023b,shridhar2023perceiver}, we explore whether existing pretrained foundation models can reason about unfamiliar objects and infer meaningful grasps without collecting any task-specific training data or performing additional training. This direction enables more autonomous and flexible robot behavior, where high-level user commands can be interpreted without detailed object specifications, and the robot can independently identify and manipulate suitable objects. To focus on this capability, we limit our study to scenes containing a single unfamiliar object, allowing us to isolate the model’s ability to perform semantic and spatial reasoning. Additionally, we aim to assess how well LMMs can localize functional regions and support spatially grounded grasp decisions in the absence of dense supervision or structured priors.

In this paper, we explore the capability of LMMs to infer feasible, task-relevant grasps from single images. We introduce \textit{ORACLE-Grasp}, a novel grasping framework (Figure \ref{fig:intro}) that treats an LMM as a semantic oracle to guide grasp selection. Rather than relying on low-level image features or geometric heuristics, ORACLE-Grasp reformulates the grasping problem as an iterative, prompt-driven reasoning process, enabling human-like, task-aware grasp predictions even for novel and amorphous objects. Our approach requires no data collection, dataset curation or pretraining, operating in a fully zero-shot manner by leveraging an existing, general-purpose LMM. Through a dual-prompt approach, ORACLE-Grasp extracts both high-level scene context and localized candidate regions, framing grasp selection as a structured, context-driven decision rather than free-form generation tasks.

The prompts are designed to be general and autonomous, allowing ORACLE-Grasp to operate without human intervention at inference time, in contrast to prior approaches that require human guidance through language commands. Critically, the use of chain-of-thought \cite{Wei2022} reasoning initiated by the first prompt encourages the model to produce affordance-aligned, human-like grasps that align with intuitive manipulation strategies. Although ORACLE-Grasp does not explicitly optimize for grasp stability, experimental results demonstrate that the predicted grasps are physically feasible and lead to high robot pick-up success rates. Additionally, the framework integrates an early stopping mechanism to improve efficiency and a depth-based refinement step to enhance grasp reliability when depth information is available. 

To summarize, our main contributions are as follows:
\begin{itemize}
    \item We introduce ORACLE-Grasp, a novel zero-shot framework for affordance-aligned grasping that leverages an LMM as a semantic oracle to infer human-like grasps without requiring training, grasp labels or custom encoders.
    
    \item We propose a dual-prompting strategy that enables structured, chain-of-thought reasoning about object function and grasp region suitability directly from a single RGB image, without external task or object grounding.
    
    \item We explore the ability of LMMs to perform context-free grasping, isolating their capacity to reason about unfamiliar, single-object scenes without requiring user-provided task annotations or structured priors.
    
    \item We show that ORACLE-Grasp supports both semantic and spatial reasoning using only pretrained models, and demonstrate how depth data can be optionally integrated to refine and improve physical grasp execution.

    \item We validate our approach in real-world robotic experiments across a variety of novel, untrained objects, demonstrating robust task-relevant grasping in fully zero-shot settings.
\end{itemize}

\section{Related Work}
\label{sec:relatedwork}

\subsection{Large Language and Multimodal Models}

LLMs have significantly advanced robotics by enabling natural language understanding for task planning, human-robot interaction and high-level decision-making \cite{ahn2022doasido, shridhar2023cliport}. These models allow robots to interpret abstract instructions (e.g., "organize the desk") and generate structured action sequences with explainable reasoning. Building on these capabilities, recent developments have introduced LMMs, which integrate vision and language processing to tackle embodied tasks \cite{brohan2023rt2, driess2023palme}. By jointly reasoning over visual inputs and textual prompts, LMMs extend the functionality of LLMs, enabling robots to perceive, interpret and act within their environments with greater semantic understanding.

Prominent examples of LMMs include GPT-4o \cite{openai2024gpt4o}, Flamingo \cite{Alayrac2022} and LLaVA \cite{liu2023llava}, which have demonstrated impressive capabilities in grounding visual scenes through natural language reasoning \cite{wang2024}. These models are typically trained on large-scale datasets comprising paired image and text data, allowing them to align visual and linguistic representations within a shared semantic space. As a result, LMMs can generalize across domains, leverage contextual information, and infer high-level semantic relationships that are critical for flexible and task-driven robotic behaviors \cite{Zhang2024}. Recently, LMMs have begun to be applied in robotics to bridge perception, reasoning and action \cite{brohan2023rt2, long2024,MonWilliams2025}, enabling new capabilities such as instruction following, semantic scene understanding and object-centric manipulation in real-world environments \cite{singh2023progprompt, lykov2024}. By leveraging their ability to reason jointly over visual and linguistic modalities, LMMs offer a promising path toward more adaptable, scalable and intuitive robotic systems that can operate with minimal task-specific training and additional data collection.


\subsection{Object and Task Grounding}

In robotic systems, \textit{grounding} refers to the process of associating abstract symbols, such as natural language commands, with perceptual representations and executable actions. Two critical forms of grounding are \textit{object grounding}, linking linguistic or symbolic references to specific objects in the environment, and \textit{task grounding}, connecting high-level task descriptions to robot actions or action plans. Early work in object grounding focused on mapping referring expressions to visual objects using either handcrafted features or statistical models \cite{tellex2011understanding}. With the rise of deep learning and large-scale vision-language models, more recent methods have enabled open-vocabulary grounding using models like CLIP \cite{Radford2021} and OWL-ViT \cite{minderer2022simple}, which align visual and linguistic representations in a shared embedding space. These models enable zero-shot recognition and grounding of previously unseen object categories based on semantic similarity.

Task grounding goes a step further by interpreting full instructions and mapping them to parameterized actions or policies. Frameworks such as SAYCAN \cite{ahn2022doasido} and PaLM-E \cite{driess2023palme} use LLMs to parse user intent and compose low-level skills or motion plans accordingly. These approaches demonstrate the ability to decompose tasks like "clean the table" into structured action sequences and ground them in sensor inputs. More recent work explores grounding in embodied agents through multimodal interaction \cite{shridhar2023perceiver} and active learning \cite{singh2023progprompt}, which enables context-aware task understanding and execution in dynamic environments. 

Despite these advances, grounding in open-world settings remains challenging due to ambiguity in language, sensor noise and the combinatorial complexity of real-world object-task-action relationships. A key limitation lies in the lack of fine-grained spatial reasoning where many models can associate text with object categories but struggle to localize specific object parts or understand spatial configurations relevant to physical interaction (e.g., “grasp the handle” vs. “pour from the spout”). This impairs the ability of robots to ground language in actionable spatial representations, especially when operating with partial observations, occlusions, or novel geometries. Ongoing research continues to explore how these models can bridge this gap by integrating semantic reasoning with geometric understanding in a generalizable and interpretable manner.


\subsection{Using LLM and LMM for Robot Grasping}

Several recent approaches leverage LLMs and LMMs for robot grasping. GraspCLIP \cite{Tang2023} enables task-oriented grasping by combining user verbal input (object and task) with visual input for grounding. However, it requires custom task-labeled datasets, limiting scalability and zero-shot generalization, and it is unclear whether it uses depth data for spatial reasoning. Unlike GraspCLIP, our proposed ORACLE-Grasp approach is domain-agnostic. Similarly, GraspGPT \cite{Tang2023b} is a task-oriented grasping framework that uses LLMs to generalize to novel object-task pairs by generating semantic descriptions from natural language instructions. While effective, GraspGPT relies on language input and learned grasp priors at inference, making it less suitable for autonomous, vision-only, zero-shot grasping compared to ORACLE-Grasp. Additionally, GraspGPT evaluates grasp candidates using learned scoring, which might prioritize grasp success over task intent.

More recently, Grasp-Anything++ \cite{Vuong2024} was presented as a large-scale language-driven grasp detection framework utilizing a diffusion model trained on over 10 million synthetic language-annotated grasp samples. It enables zero-shot performance guided by natural language instructions, however, its reliance on extensive training data and language input at inference time constrains its autonomy. Unlike ORACLE-Grasp, it lacks interpretable spatial reasoning and geometric refinement. LAN-grasp \cite{Mirjalili2024} is a zero-shot semantic grasping method that integrates an LLM, LMM and a conventional grasp planner. The LLM identifies the appropriate object part from a user-provided label, which the LMM grounds visually, and a grasp planner generates constrained grasps. However, LAN-grasp's dependence on external object labels and full 3D reconstruction limits its autonomy and real-time applicability.

ShapeGrasp is a zero-shot task-oriented grasping method that decomposes novel objects into simple geometric parts, represented as graphs \cite{Li2024ShapeGrasp}. An LLM assigns semantics and selects the most suitable part to grasp based on the object name and task. The grasp is then executed on the part deemed most suitable, using basic heuristics over the part’s 3D geometry. However, ShapeGrasp relies heavily on handcrafted geometric decomposition, segmentation and symbolic reasoning, limiting its adaptability to complex objects. It lacks direct visual grounding and spatial precision, as it does not integrate LMM or learn from dense perceptual data. Unlike our approach, which performs end-to-end grounding and learns spatially-informed affordance-aligned grasps, ShapeGrasp depends on manual heuristics and coarse geometry, reducing its generality and robustness in cluttered or ambiguous scenes.

FoundationGrasp \cite{Tang2025} is another task-oriented grasping framework that utilizes LLMs and VLMs for generalization to novel objects. It combines semantic and geometric reasoning through LLM descriptions, LLM/VLM-based encoders and a Transformer for grasp evaluation. However, FoundationGrasp requires task-specific user input and learned grasp evaluation networks, which restricts its autonomy and zero-shot capabilities. Unlike ORACLE-Grasp, it relies on supervised components and lacks interpretable, iterative spatial reasoning, potentially reducing its generalizability. Since LMMs typically lack precise spatial reasoning, these approaches often rely on custom-trained grasp detection models, which constrain their ability to generalize to new objects and tasks.

\section{Methods}

While state-of-the-art LMM offer powerful semantic understanding and zero-shot object recognition, they fall short when applied to fine-grained part localization for robotic grasping. Outputs such as bounding boxes and masks often lack the spatial precision to isolate specific graspable parts, show inconsistency in the exact region they highlight across similar inputs, and do not convey the orientation required for proper gripper alignment. For instance, a handle might be detected with a bounding box that is loosely aligned, oversized or arbitrarily rotated, making it unclear how the gripper should be positioned. Even segmentation masks, while richer in detail than boxes, often yield imprecise or fragmented regions that shift with minor input changes, offering no consistent contour or graspable axis. These limitations undermine the reliability of grasp execution, where even small errors in location or angle can lead to failure. To address this, this section introduces our method, which restructures grasp localization as an iterative decision process, enabling accurate and robust grasp prediction without explicit training.


\subsection{Problem Definition}

We address the task of affordance-aligned grasp prediction for unfamiliar objects using a robotic arm equipped with an RGB-D camera. The camera may be mounted on the robot’s end-effector or statically placed in the workspace, assuming proper calibration. We assume a simplified setting where a single, previously unseen object is present in the robot's workspace. This setup allows us to isolate and study the semantic and spatial reasoning capabilities of LMMs. In future work, this assumption could be relaxed by incorporating context-aware reasoning based on the task or user-provided verbal instructions. It is important to note that a standard RGB camera can be used in place of an RGB-D camera; however, this excludes the possibility of depth-based grasp refinement, which is discussed in a later section. Compensating for the lack of depth data would require additional techniques, such as visual servoing, which are beyond the scope of this work.

The input to the system is an RGB image $\ve{I}_i$ and a corresponding depth map $\ve{D}_i$ of the scene, both captured from a single viewpoint. The object of interest is assumed to be visible in the image, without requiring prior segmentation or labeling. The system predicts a single affordance-aligned grasp pose for the object, defined as a grasp tuple $(\bve{p}_i, \theta_i)$, where $\bve{p}_i \in \mathbb{N}^2$ is the grasp position and $\theta_i \in \mathbb{R}$ is the required gripper orientation angle, both with respect to the image coordinate frame. This grasp should not only be physically feasible but also semantically meaningful, aligning with the object’s most likely functional usage (e.g., grasping a knife by the handle rather than the blade). The goal is to perform this prediction in a zero-shot setting: no object labels, segmentation masks, training data or prior knowledge about object geometry or category are provided.



\subsection{Grasp Candidate Generation}

LMMs, while proficient at generating natural language descriptions and answering image-related queries, exhibit limitations in spatial precision when embedding images into their latent space. This deficiency hinders their ability to produce pixel-accurate oriented bounding boxes for specific object parts, such as \texttt{'the handle of the mug'} or \texttt{'the handle of the drill'}. Direct bounding box generation requests often yield imprecise or inconsistent outputs. Hence, we propose a novel strategy within the ORACLE-Grasp framework to guide the LMM. We first define the image tiling operation $\Gamma_{a \times b}(\ve{I}_j)$ as the process of discretizing the image $\ve{I}_j$ into an $a \times b$ grid of equally sized tiles. In doing so, the problem becomes a multiple–choice question of candidate regions, where the LMM can select a tile instead of a free–form generation task. This approach is particularly effective for objects without obvious handles (e.g., a cardboard, a wallet or a shoe) and leads to more human–like grasp selections.

\begin{table*}
    \centering
    \caption{The two prompts used to provide context and locate a region to grasp}
    \label{tb:prompts}
    \begin{tabular}{p{0.15\linewidth}  p{0.85\linewidth}}\toprule
        (a) Scene Context Prompt (SCP) & \texttt{Please provide a short, concise description of the principal object present in the image, focusing on the parts of the object that could be used for grasping. If the object has a clear handle or grip, mention that; if not, describe a cylindrical or otherwise ergonomically graspable section of the object. Avoid extraneous details unrelated to how one might physically grasp the object. Keep your answer to one sentence.} \\
        & \\\midrule
                
        (b) Grasp Region Prompt (GRP) & 
        \texttt{Based on the following image context: \textit{<SCP response>}, analyze the provided image and determine the optimal grid cell, from (0,0) to (columns, rows - 1), that corresponds to the best grasping area for the object. Focus exclusively on the object (ignore all background and surrounding elements).} \\
        & \texttt{CONSIDER THE FOLLOWING:} \\
        & \texttt{1. Prioritize areas that resemble handles or have handle-like features.}\\
        & \texttt{2. If no handle is present, select the most stable area.}\\
        & \texttt{3. Avoid areas that could interfere with the object's functionality.}\\
        & \texttt{IMPORTANT:} \\
        & \texttt{1. Your response MUST follow exactly the format below.}\\
        & \texttt{2. DO NOT include any additional text, markdown formatting, or commentary.}\\
        & \texttt{OUTPUT FORMAT:} \\
        & \texttt{GRID\_CELL: <cell\_number>}\\
        & \texttt{EXPLANATION: <brief explanation of your choice>}\\
        \bottomrule
    \end{tabular}
\end{table*}
\begin{figure*}
    \centering
    \includegraphics[width=\linewidth]{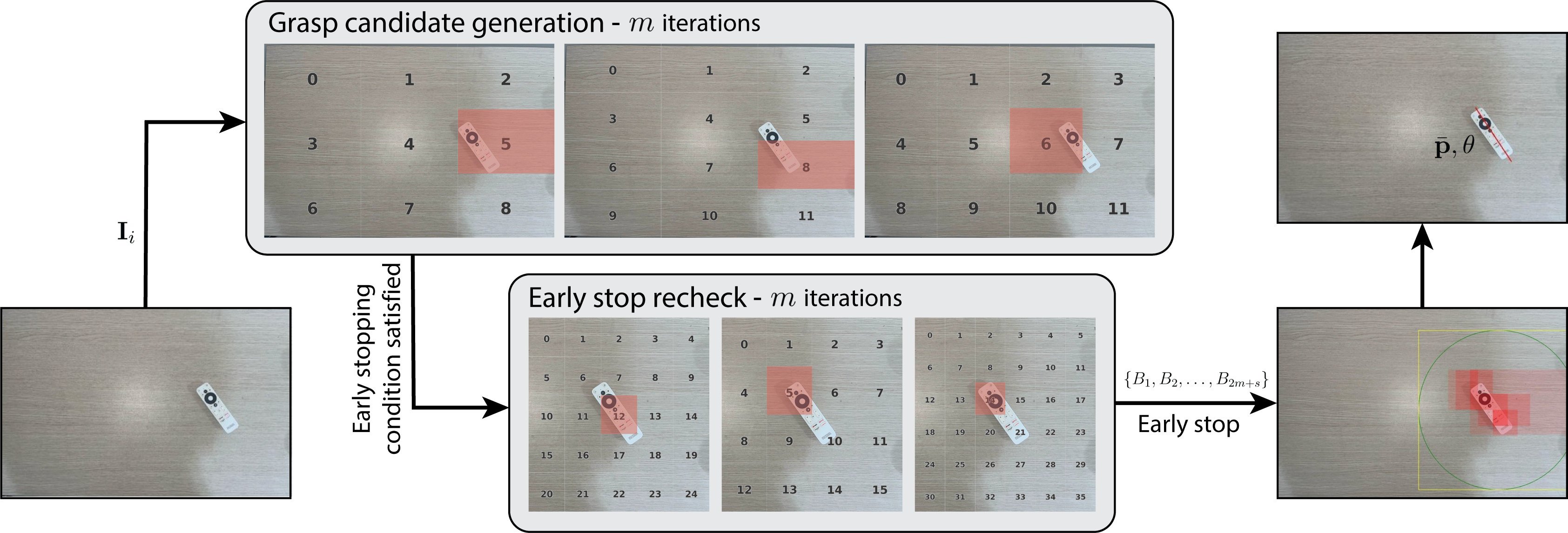}
    \caption{Illustration of a grasp candidate generation given an image $\ve{I}_i$ of a remote control object. Red cells on the grid denote selection for grasping by the LMM. After $m=3$ iterations with the GRP prompt over different grid configurations, stopping criterion \eqref{eq:condition} is satisfied leading to an additional $m$ iterations on a cropped image. Finally, the position $\bve{p}$ and orientation angle $\theta$ of the required grasp are outputted. }
    \label{fig:scheme_stop}
\end{figure*}

\textbf{Dual-prompt masking.} To guide the LMM, we propose the use of a dual-prompt process that leverages tool calling capabilities. The first prompt, referred to as the Scene Context Prompt (SCP), is paired with the original image $\ve{I}_i$ and instructs the model to generate a concise description of the primary object in the scene, with an emphasis on grasp-relevant features. Here, the SCP acts as a tool call that extracts high-level semantic context, helping the model to reason more effectively about where and how to grasp the object. An example of this prompt is shown in Table \ref{tb:prompts}.a. This step encourages the model to perform Chain-of-Thought (CoT) reasoning about the scene, generating a semantic context that guides the following prompt's spatial selection.
The second prompt, termed Grasp Region Prompt (GRP) and seen in Table \ref{tb:prompts}.b, aims to approximate a grasp location. To overcome the spatial imprecision of LMMs and facilitate effective reasoning over grasp locations, we discretize the image space into a set of discrete candidate areas. It is rolled out in multiple iterations to acquire a probabilistic estimation of the grasp region. In each iteration, the GRP is paired with an image $\Gamma_{u \times v}(\ve{I}_i)$ such that $u,v\in\{3,4,\dots,9\}$. The values for $v$ and $u$ are selected without repetition, starting with coarse grids and gradually refined through additional GRP queries. 

The GRP incorporates the contextual output of the SCP (i.e., \texttt{<SCP response>}) to instruct the LMM to choose the optimal tile from the discretized image (Figure \ref{fig:scheme_stop}). Crucially, the GRP output includes a required \texttt{<brief explanation of your choice> field, which serves as a mechanism for CoT reasoning (Table }\ref{tb:prompts}.b). By forcing the LMM to articulate its semantic and functional rationale before committing to a specific grid cell, the framework ensures the model actively applies contextual constraints to the spatial domain, such as prioritizing handles or avoiding fragile parts. This requirement acts as a vital bridge between the high-level affordances identified by the SCP and the low-level spatial grid, grounding the final grasp selection in a semantically-informed reasoning process rather than a purely geometric estimation. Furthermore, these explanations provide interpretability, allowing the zero-shot decision process to be verified against the object's perceived functional properties. Repeating the GRP procedure over $K$ iterations with varying grid configurations produces a set of candidate regions $\mathcal{B}=\{B_1, B_2, \dots, B_K\}$, where $B_j$ denotes a rectangular pixel mask over the image, selected by the LMM during iteration $j$. Discretizing the image space into a grid of candidate areas enables a more robust and simple selection of grasp regions.

\textbf{Grasp position and orientation.} For each pair of overlapping rectangular masks $B_j \in \mathcal{B}$ and $B_k \in \mathcal{B}$, we compute their Intersection-over-Union (IoU), denoted as $C_{ik}=IoU(B_j, B_k)$. If $C_{ik} > \gamma$, for a predefined threshold $\gamma$, the intersection region $B_j \cap B_k$ is added to the $\mathcal{B}$. A higher IoU value suggests a more promising grasp region, prompting the inclusion of an additional rectangle that reinforces areas with high grasp potential. After $K$ iterations and the inclusion of $s$ overlapping regions, the set $\mathcal{B}$ contains a total of $K + s$ rectangular pixel masks. In the subsequent step, the center coordinate $\ve{p}_j \in \mathbb{N}^2$ of each mask $B_j \in \mathcal{B}$ is computed and collected into a set $\mathcal{M}$, representing the centers of all masks. The centroid $\bve{p}$ of set $\mathcal{M}=\{\ve{p}_1,\ldots,\ve{p}_{K+s}\}$ determines the final grasp point. Furthermore, Principal Component Analysis (PCA) is applied to the set of center points in $\mathcal{M}$ to extract the principal eigenvector $\ve{v}^*=(v^*_x,v^*_y)$ associated with the largest eigenvalue. The grasp orientation angle $\theta$ can be extracted from
\begin{equation}
    \theta=\arctantwo(v^*_y,v^*_x).
\end{equation}
Consequently, we acquire the required grasp position $\bve{p}$ and orientation $\theta$ with respect to the image's coordinate frame.

\textbf{Early stopping criterion.} The above process is run for $K$ iterations by default. To reduce processing time and to mitigate the potential effect of multiple grasping regions, 
we incorporate an early stopping mechanism which also enforces focus on a specific part of the object. This mechanism assesses the spatial distribution of the rectangular mask centers as an indicator of convergence. After $m < K$ iterations, the centroid $\bar{\ve{p}}$ of the current set of center points ${\ve{p}_1, \ldots, \ve{p}_m}$ is computed. Let $0 < \rho < 1$ be a predefined factor, and let $D$ denote the image diagonal. 
If the maximum distance from any center point to the centroid satisfies
\begin{equation}
    \label{eq:condition}
    \max_{i\in\{1,\ldots,m\}}\|\ve{p}_i-\bve{p}\|<\rho D,
\end{equation}
an additional $m$ iterations are performed on a cropped region of the image centered around the object. The cropping bounds are defined based on the largest diagonal that fits within the union of the $m$ selected grid cells ${B_1, \ldots, B_m}$. If condition \eqref{eq:condition} is satisfied after these additional iterations, the process terminates early. Otherwise, if the condition is never met, the procedure defaults to selecting the acquired solution after completing all $K$ iterations. If early stopping occurs, $s$ overlapping regions are calculated with existing masks using the IoU method. Consequently, set $\mathcal{B}$ comprises $2m+s$ rectangular masks, which provide the centroids and eigenvectors used to determine grasp position and orientation. This early stopping strategy not only accelerates the overall pipeline execution but also enhances robustness by ensuring that only consistent and reliable grasp candidates contribute to the final decision.

\textbf{Orientation refinement.} Arbitrary rotation of the object’s graspable region relative to the image axes causes grid misalignment, leading to inadequate coverage and suboptimal mask fitting.
This misalignment can significantly affect the accuracy of the process, particularly when the object spans a large portion of the image. To address this, we utilize PCA on the set of center points to estimate their dominant orientation. If the resulting orientation angle $\alpha$ exceeds a predefined threshold $\alpha_{min}$, the image is rotated by $-\alpha$ to better align the object with the grid structure. A new set of $m$ iterations is then executed. The resulting pixel masks are subsequently transformed back into the original image coordinate system via inverse rotation by $\alpha$. This refinement improves the estimation of both the grasp point and its orientation, enhancing the system's precision and robustness in challenging edge cases.

The precision of the orientation estimation $\theta$ is derived from the spatial distribution of the selected grid centers. For elongated objects, the set of selected cells $\mathcal{B}_j$ forms a point cloud where the variance along the object's principal axis significantly outweighs the transverse variance. This results in a dominant principal eigenvector, ensuring directional precision even with coarse discretization. To handle non-linear or curved geometries, the iterative refinement process crops and rescales the image around the region of interest. This zooming mechanism effectively performs a local linearization of the object's geometry, allowing the PCA to compute an accurate grasp axis on a localized, nearly-linear segment of the object. This combination of global reasoning and local geometric analysis allows ORACLE-Grasp to maintain high directional accuracy across diverse object profiles.


\subsection{Grasp Position Refinement}
\label{sec:adjust}

\begin{figure}
    \centering
    \includegraphics[width=\linewidth]{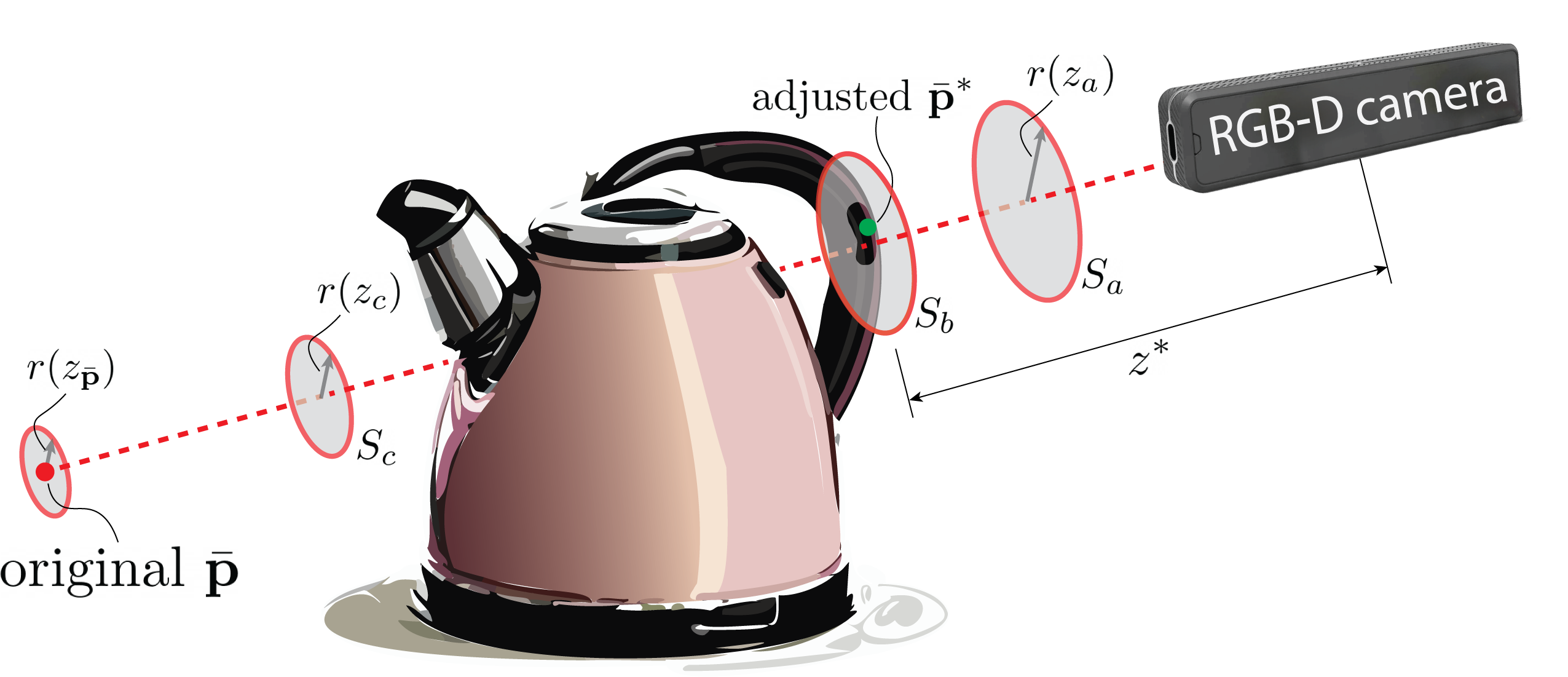}
    \caption{Illustration of a case where the predicted grasp position (circular red marker) $\bve{p}$ slightly missed the kettle's handle in the RGB image, leading to wrong depth estimation $z_{\bve{p}}$. Circular regions (such as $S_a$ and $S_c$ with radii $r_a$ and $r_c$, respectively) along the line connecting the camera and original $\bve{p}$ are checked for improved depth estimation toward refined grasp position $\bve{p}^*$ (circular green marker) at depth $z^*$.}
    \label{fig:adjustment}
\end{figure}

The above process considers approximation of the required grasp point using RGB planar perception alone. Also, it uses several grid approximations that cannot be thoroughly refined due to a long computational time. Hence, the resulting grasp point may slightly deviate from the object, pointing instead to empty space rather than the solid structure of the object. For example, when the grasp pipeline predicts a position near the center of a kettle’s handle, which is often concave and semi-circular as illustrated in Figure \ref{fig:adjustment}, the depth sensor may incorrectly report an exaggerated depth value due to capturing the background through the hollow region. 

To ensure a safe and accurate grasp, we employ a heuristic depth-based grasp refinement procedure that evaluates the local geometry around the predicted grasp point. Using the depth image $\ve{D}_i$, the system projects the expected closing area of the gripper based on the depth $z_{\bve{p}}$ at the predicted grasp point $\bve{p}$. Let $r_{ee}$ be the required clearance radius of the robot's end-effector, and $f$ be the the camera’s focal length, the clearance radius in image units (i.e., in pixels) at depth $z$ is computed by
\begin{equation}
    r(z)=\frac{f\cdot r_{ee}}{z}.
\end{equation}
This radius determines the region that the gripper fingers will occupy in the image, scaled inversely with depth. 

To verify the feasibility of the grasp, we sample $w$ depth values $\mathcal{Z} = \{ z_j \}_{j=1}^w$, where each value is defined by $z_j=\frac{j}{w} \cdot z_{\bve{p}}$. For each $z_j \in \mathcal{Z}$, we define a circular region $S_j \subset \ve{D}_i$ in the depth image $\ve{D}_i$, centered at the original grasp point $\bve{p}$ and with radius $r(z_j)$. Within each region $S_j$, we extract the minimal depth value $\tilde{z}_j \in S_j$, along with its corresponding pixel location $\tve{p}_j \in \mathbb{N}^2$. The final refined grasp position is then defined as $\bve{p}^* = \tve{p}_t$, where
\begin{equation}
    t = \argmin_{j \in \{1, \ldots, w\}} \tilde{z}_j,
\end{equation}
and the corresponding depth is $z^* = z_t \in S_t$. This procedure ensures that the grasp targets the closest valid surface within the line connecting the camera and the original outputted position $\bve{p}$. Hence, the robot would target a physically valid and reachable region, avoiding misleading grasp locations due to insufficient grid refinements. 

\label{sec:method}

\section{Experiments}
\label{sec:experiments}

In this section, we evaluate the performance of ORACLE-Grasp in predicting human-like grasps, and enabling robotic pick-up. For our experimental validation, we adopt LLaMa 3.2-Vision (MLLaMA) \cite{Grattafiori2024} as the primary LMM. This choice was driven by the objective of utilizing state-of-the-art open-weight models to ensure the framework remains accessible, reproducible, and deployable on local hardware without reliance on proprietary APIs. While larger closed-source models (such as GPT-4o) consistently achieve higher scores on standard multimodal benchmarks, we utilize the 11B-parameter LLaMa model as a reliable baseline. Demonstrating the effectiveness of ORACLE-Grasp with this more compact and open-source backbone suggests that the framework’s performance would likely be further enhanced if integrated with more powerful, high-parameter proprietary models.

The MLLaMA model was deployed on a GPU cloud instance equipped with an NVIDIA H200 SXM GPU, 24 CPUs, and 377 GB of RAM. Though cloud inference is faster, preliminary tests show ORACLE-Grasp is also deployable on local hardware, with runtime as the main trade-off. Following extensive hyperparameter tuning, we set the key parameters as follows: number of GRP iterations $K = 6$, IoU threshold $\gamma = 0.4$, early stopping window $m = 3$, and stopping factor $\rho = 0.3$.

\begin{table}[]
\caption{Position and orientation errors for ORACLE-Grasp over test-set \textit{I}}
\label{tb:eval}
\centering
\begin{tabular}{lcc}\toprule
\multicolumn{1}{c}{\multirow{2}{*}{Prompt variation}} & Position  & Orientation    \\
\multicolumn{1}{c}{}           & NRMSE ($\times10^2$) & MAE ($^\circ)$ \\\midrule
Baseline prompt         & 61 $\pm$ 50 & - \\   
ORACLE-Grasp w/o SCP         & 6.8 $\pm$ 6.1 & 24.3 $\pm$ 20.2 \\
ORACLE-Grasp w/ SCP and GRP  & \cellcolor[HTML]{C0C0C0}5.1 $\pm$ 4.2 & \cellcolor[HTML]{C0C0C0}19.0 $\pm$ 14.8  \\
\bottomrule
\end{tabular}
\end{table}


\subsection{Test-sets}

To evaluate the generalization capability of ORACLE-Grasp, we design two test-sets composed entirely of novel objects that have not been seen by the model during pretraining. The ability to successfully predict appropriate grasp points on such unfamiliar objects serves as a direct measure of the model’s capacity to generalize beyond fixed training distributions, a key limitation of traditional supervised grasping methods. Test-set \textit{I} consists of 100 RGB images featuring diverse objects in varied settings. Of these, 90 were collected from the web, offering a broad range of object types and environments. The remaining 10 images were generated using DALL$\cdot$E, specifically prompted to depict unfamiliar objects in realistic yet imaginative scenes. This AI-generated subset is deliberately included to challenge both human and model reasoning, pushing the boundaries of generalization to unfamiliar and abstract object forms. Test-set \textit{II} includes 20 RGB-D images captured using a ZED-Mini camera. These images contain real-world objects such as a brush, toothpaste tube and carton box. They are used to assess grasp prediction in realistic settings and enable refinement using depth data. By evaluating performance on these unseen and diverse objects without additional training, we aim to demonstrate ORACLE-Grasp’s ability to generalize affordance-aligned grasping behavior in true zero-shot settings.



\subsection{Evaluation metric}

The ORACLE-Grasp framework produces grasps guided by the reasoning capabilities of an LMM. Therefore, evaluating the predicted grasps should involve comparing them to human judgments, which typically align with the object's affordances. For this purpose, each image in the test-sets is annotated by an uninvolved human participant. These annotations were collected using a dedicated Graphical User Interface, allowing annotators to click on potential grasp locations and draw lines representing the intended grasp direction. The participant was instructed to subjectively annotate potential grasp centers along with the corresponding orientations, reflecting natural hand alignments for each grasp. Multiple annotations were allowed to capture different task-dependent grasp strategies - for example, a drill could be grasped by its handle for a drilling task, or by its cylindrical body when intended for handover. This process yields human annotated grasp positions $\{\bve{p}_{an,1},\bve{p}_{an,2},\ldots\}$ (in pixel units) and orientations $\{\theta_{an,1},\theta_{an,2},\ldots\}$ for each image in the test-sets, denoting human-like and affordance-aligned grasps. 

The predicted grasps by ORACLE-Grasp are evaluated against human annotations. In test-set \textit{I}, where no metrical data is available and object scales vary, we compute the Normalized Root Mean Squared Error (NRMSE) of the grasp position by dividing the planar distance between the predicted position $\bve{p}$ and closest annotated position $\bve{p}_{an,j}$ by the diameter $d$ of the object’s bounding circle: $e = \frac{|\bve{p} - \bve{p}_{an,j}|}{d}$. This yields a unit-less, scale-invariant measure of positional accuracy. For test-set \textit{II}, which includes depth data, we report standard RMSE in metric units. In both test-sets, orientation accuracy is evaluated using the Mean Absolute Error (MAE) between the predicted angle $\theta$ and annotated angle $\theta_{an,j}$ across all images.

\begin{figure}
    \centering
    \includegraphics[width=\linewidth]{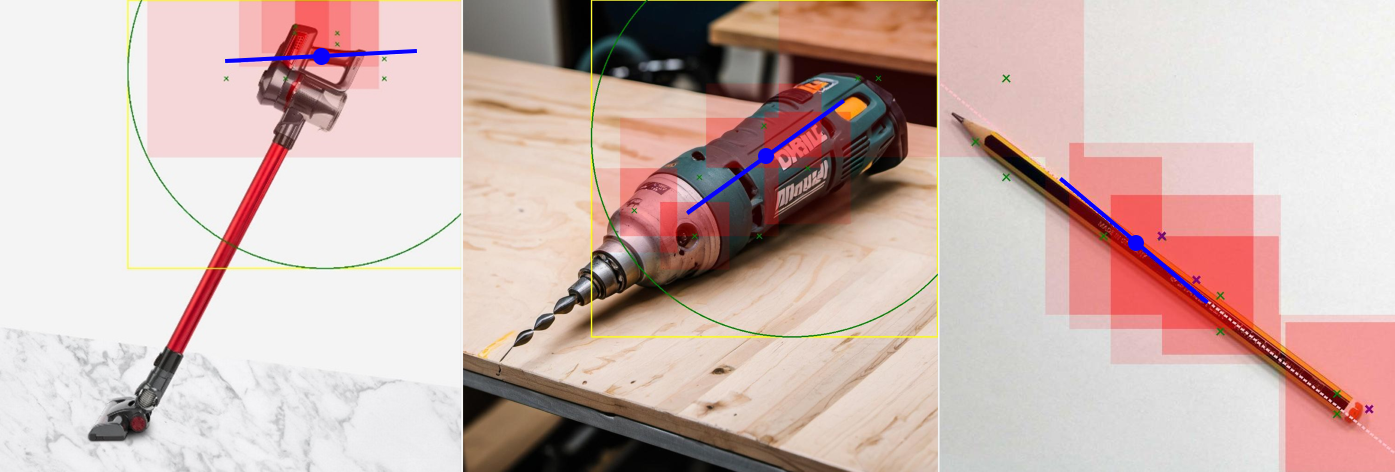}
    \caption{Predicted grasps on three RGB images from test-set \textit{I}, illustrating the predicted grasp position (blue dot) and orientation (blue line). Objects include, from left to right: a vacuum cleaner, an AI-generated drill and a pencil. Red cells indicate regions selected by the LMM as potential grasping areas.}
    \label{fig:objects_1}
\end{figure}
\begin{figure*}
    \centering
    \includegraphics[width=\linewidth]{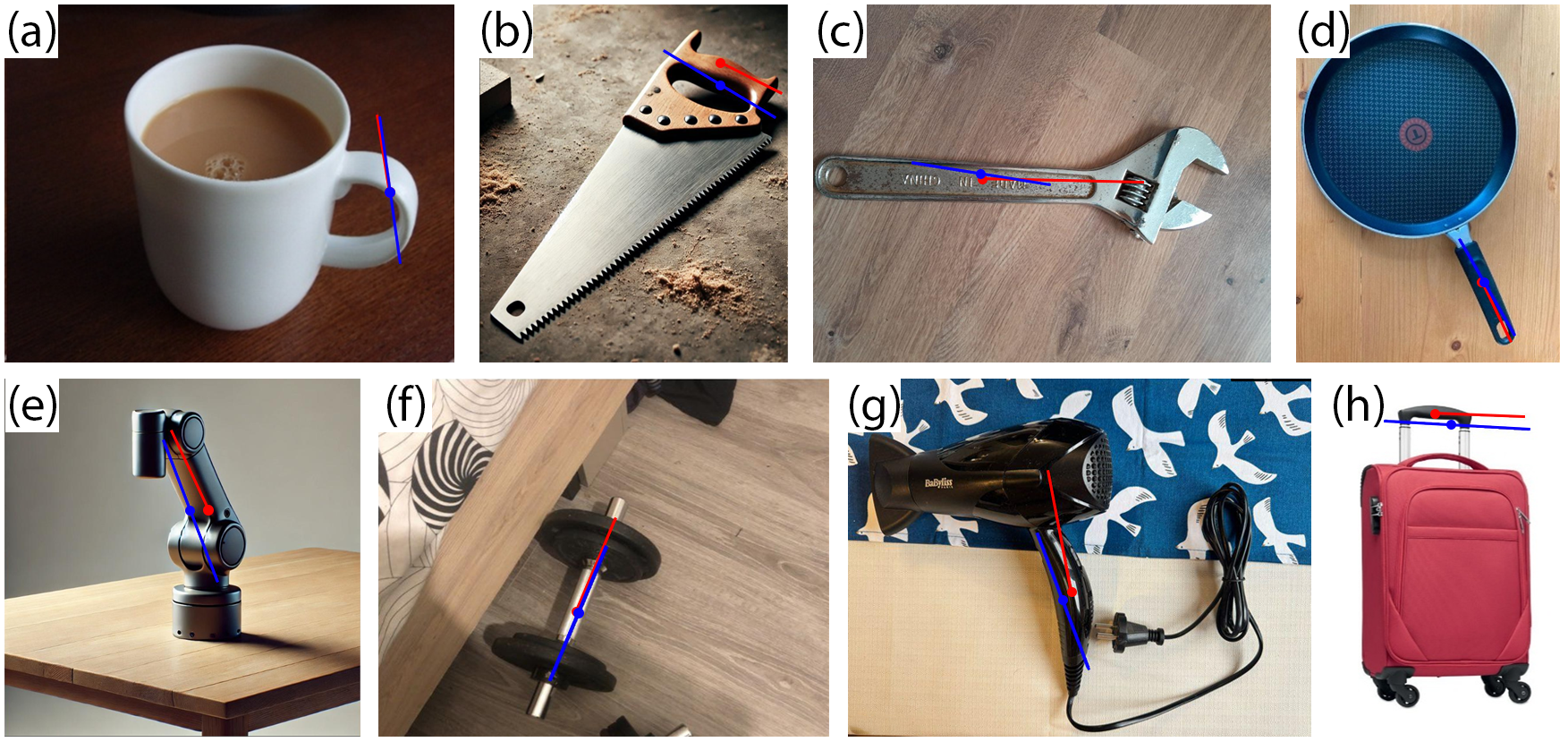}
    \caption{Predicted grasps (blue) versus human annotations (red) on RGB images from test-set \textit{I}. Objects shown with their respective NRMSE: (a) Mug – 0.0385, (b) Saw – 0.027, (c) Adjustable wrench – 0.026, (d) Frying pan – 0.020, (e) Robot arm – 0.086, (f) Dumbbell – 0.074, (g) Hair dryer – 0.018, and (h) Suitcase – 0.046.}
    \label{fig:testset1}
\end{figure*}

\begin{figure}
    \centering
    \includegraphics[width=\linewidth]{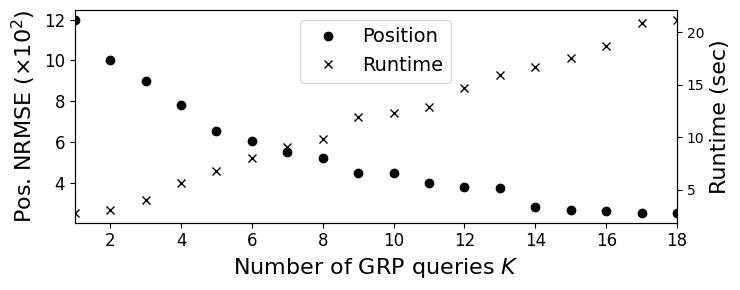}
    \caption{Position NRMSE and computation runtime with regard to the number of GRP queries $K$ exerted in ORACLE-Grasp.}
    \label{fig:error_time_vs_GRP}
\end{figure}
\begin{table}[ht]
\centering
\footnotesize
\caption{Grasp prediction accuracy and pick-up success rates for test-set \textit{II} (Ablation Study)}
\label{tb:objects}
\setlength{\tabcolsep}{2.5pt} 
\begin{tabular}{lccccccc}\toprule
\multirow{2}{*}{Object} & \multicolumn{4}{c}{Position RMSE (mm)} & \multicolumn{2}{c}{Orientation MAE ($^\circ$)} & \multirow{3}{*}{\begin{tabular}[c]{@{}c@{}}Success rate \\(\%)\end{tabular}} \\\cmidrule(lr){2-5} \cmidrule(lr){6-7}
& \multirow{2}{*}{w/o GR} & \multicolumn{5}{c}{w/ GR} & \\\cmidrule(lr){3-7}
 &  & w/o OR & w/o BE & Full & w/o OR & Full & \\
\midrule
Coffee bag        & 34.8  & 8.9   & 9.9  & 8.9  & 1.01  & 1.01  & 80  \\
Phone charger     & 19.91 & 8.32  & 14.7 & 4.0  & 6.50  & 1.51  & 100 \\
USB cable         & 21.49 & 25.0  & 10.9 & 25.0 & 6.29  & 6.29  & 100 \\
Brush             & 3.3   & 14.3  & 21.8 & 3.3  & 11.72 & 0.72  & 100 \\
Cardboard box     & 29.41 & 8.1   & 17.0 & 8.1  & 9.92  & 3.74  & 80  \\
Briefcase         & 30.6  & 21.0  & 32.3 & 21.0 & 8.22  & 8.22  & 60  \\
Glue gun          & 35.11 & 7.2   & 26.6 & 7.2  & 15.38 & 15.38 & 80  \\
Saw               & 17.5  & 16.88 & 8.6  & 12.3 & 19.5  & 2.6   & 80  \\
Kneading knife    & 3.0   & 10.42 & 7.8  & 3.0  & 7.84  & 7.84  & 100 \\
Toothpaste        & 8.9   & 8.9   & 10.5 & 8.9  & 5.23  & 5.23  & 100 \\
Strengthener      & 15.0  & 18.8  & 16.4 & 18.8 & 18.85 & 18.85 & 80  \\
Dustpan           & 30.6  & 20.68 & 7.8  & 11.7 & 28.92 & 7.37  & 80  \\
Rubber duck       & 22.51 & 5.2   & 32.5 & 5.2  & 2.64  & 2.64  & 100 \\
Stapler           & 10.82 & 12.81 & 13.4 & 9.2  & 17.14 & 13.54 & 100 \\
Plastic cylinder  & 29.77 & 6.5   & 11.8 & 6.5  & 13.05 & 13.05 & 80  \\
Plastic cup       & 27.58 & 16.9  & 8.4  & 16.9 & 10.72 & 10.72 & 100 \\
Desk C-Clamp      & 18.97 & 26.18 & 18.8 & 19.0 & 0.15  & 4.76  & 80  \\
Remote controller & 19.21 & 2.3   & 9.4  & 2.3  & 4.92  & 4.92  & 100 \\
Soldering holder  & 21.14 & 20.49 & 34.7 & 10.4 & 7.46  & 7.2   & 80  \\
Bubble level tool & 19.12 & 13.95 & 17.5 & 4.3  & 9.58  & 9.37  & 80  \\midrule
Total             & 20.9$\pm$9.5 & 13.6$\pm$6.9 & 16.5$\pm$8.7 & \cellcolor[HTML]{C0C0C0}10.3$\pm$6.6 & 10.3$\pm$7.0 & \cellcolor[HTML]{C0C0C0}7.2$\pm$5.0 & 88 \\
\bottomrule    
\multicolumn{8}{l}{\footnotesize{* GR - Grasp Refinement (Sec. \ref{sec:adjust})}} \\
\multicolumn{8}{l}{\footnotesize{* OR - Orientation Refinement}} \\
\multicolumn{8}{l}{\footnotesize{* BE - Brief Explanation}}
\end{tabular}
\end{table}
\begin{figure*}
    \centering
    \includegraphics[width=\linewidth]{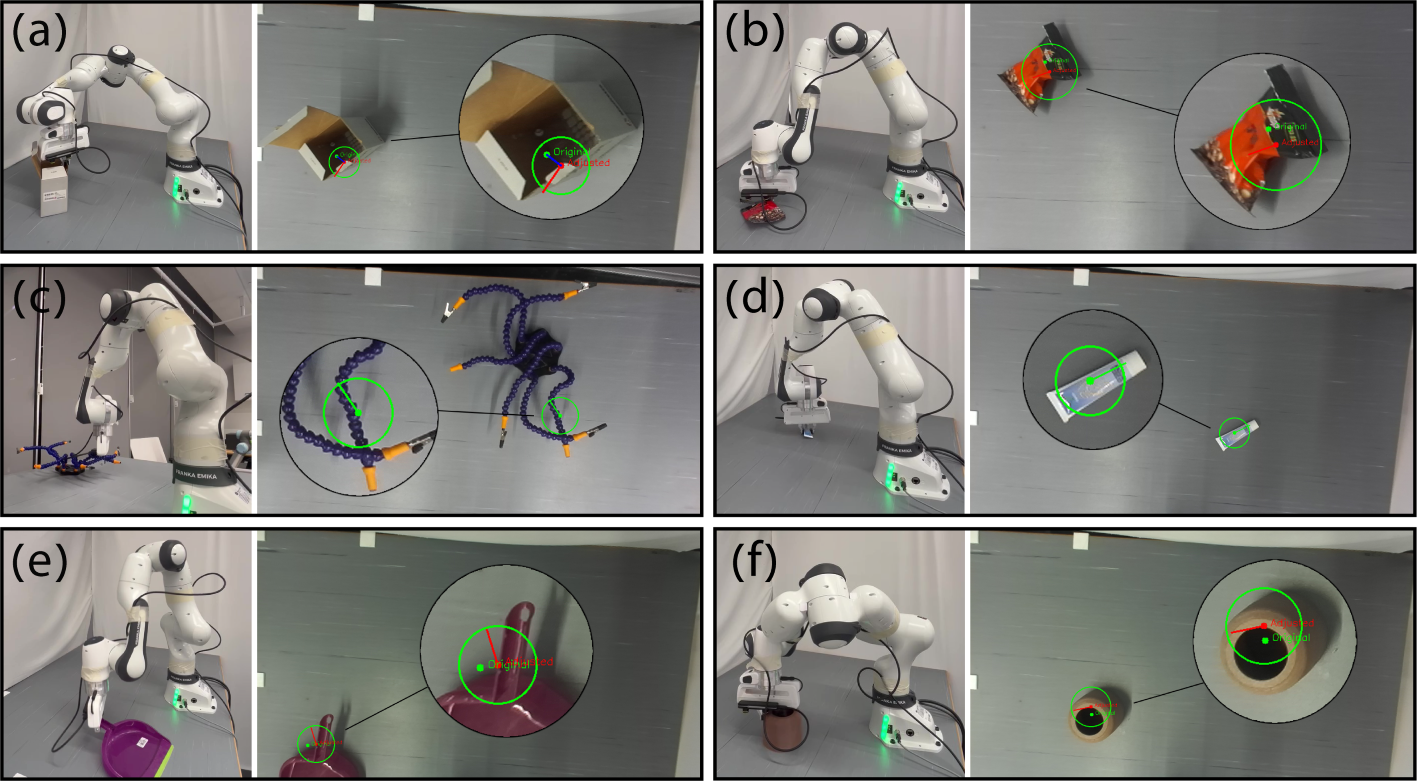}
    \caption{Real-world objects from test-set \textit{II} being grasped by a robot arm using predictions from ORACLE-Grasp. For each object, the figure shows a snapshot of the robot executing the grasp and the corresponding RGB image captured by the on-board camera. The demonstrated objects include: (a) cardboard box, (b) coffee bag, (c) six-arm soldering holder, (d) toothpaste tube, (e) dustpan, and (f) Plastic cylinder. In some cases, the initial predicted grasp (in green) was refined using depth data (final grasp shown in red), as described in Section \ref{sec:adjust}. The green circles mark the clearance region with radius $r(z_{\bve{p}})$.}
    \label{fig:grasps}
\end{figure*}


\subsection{Grasp error evaluation}

We begin by evaluating grasp accuracy against human annotations in test-set \textit{I}. We include a baseline where the LMM is prompted to return a bounding box coordinates, with the grasp position taken as the box's center. Though lacking orientation, this serves as a reference for NRMSE comparison. We also compare performance with and without the SCP. Without it, the model receives only the GRP, which lacks contextual cues. Table \ref{tb:eval} reports the NRMSE and MAE for test-set \textit{I}. The baseline results indicate that directly prompting the LMM for a bounding box yields imprecise and often irrelevant regions. Note that grasp refinement was not applied to test-set \textit{I} due to the absence of depth data, and the reported results exclude this step. The results demonstrate improved accuracy when the SCP is used alongside the GRP, compared to using the GRP alone. This highlights the significant impact of the SCP: by prompting the model to reason about the scene, it encourages context-aware, task-relevant grasp region selection. Figure \ref{fig:objects_1} demonstrates predicted grasps over three RGB images along with a heatmap representing the LMM's grid selections. Similarly, Figure \ref{fig:testset1} presents the grasp prediction by ORACLE-Grasp along with the human-annotated grasps. The predicted grasps reflect human-like reasoning, focusing on task-relevant regions such as handles. 

We further evaluate the inference time required for generating a grasp prediction. First, the duration of a single GRP query is measured as a function of grid size $u \times v$, executing 500 GRP queries over test-set \textit{I} with $u$ and $v$ sampled uniformly. The average inference time is $0.93 \pm 0.07$ seconds per query. The low standard deviation suggests that GRP inference is largely insensitive to grid resolution, allowing flexible adjustment with minimal computational cost. The average total inference time for ORACLE-Grasp to predict a complete grasp over the entire test-set is $7.73 \pm 2.05$ seconds. This latency can be further reduced with faster GPUs or using multiple ones. Notably, 89\% of ORACLE-Grasp runs terminated early due to the early stopping criterion, underscoring its practical efficiency. To illustrate the trade-off between grasp accuracy and runtime, Figure \ref{fig:error_time_vs_GRP} shows the position NRMSE and runtime on test-set \textit{I} as a function of the number of GRP queries $K$, revealing improved accuracy at the cost of increased computation.



\subsection{Robot grasping experiments}

In the following experiment, we assess the effectiveness of ORACLE-Grasp in guiding a robotic arm to grasp real-world objects. We use a Franka Research 3 robot arm equipped with a simple parallel gripper, and a ZED-Mini camera mounted at the base of the gripper. For each grasp attempt, the arm first positioned the camera to capture an overhead image of the table. ORACLE-Grasp then processed this image to predict the grasp point and orientation, which the robot subsequently used to attempt a pick-up by simply employing ROS MoveIt motion planning framework.

\newpage
The evaluation covers all 20 everyday objects from test-set \textit{II}, with five grasp trials per object. As before, a human annotator labeled the grasp points and orientations without observing the robot’s actual performance. Table \ref{tb:objects} summarizes the experimental results, detailing the position RMSE, orientation MAE, and the pick-up success rate across all trials. To evaluate the individual contributions of our system components, we present an ablation study comparing the performance of the full framework against configurations without: (i) depth-based Grasp Refinement (GR), (ii) the Brief Explanation (BE) CoT format in the GRP, and (iii) the Orientation Refinement (OR) module. The availability of depth data in test-set \textit{II} enables grasp refinement, which significantly improves positional accuracy by allowing precise adjustments before execution. Figure \ref{fig:grasps} shows snapshots from several grasping trials of the robot, displaying real-time RGB inputs alongside ORACLE-Grasp’s predicted grasps.


Experimental results underscore ORACLE-Grasp’s robust performance across a diverse range of real-world objects, highlighting significant accuracy gains from the iterative grasp refinement, orientation refinement and the CoT reasoning facilitated by the 'Brief Explanation' field in the GRP. By forcing the model to articulate its functional rationale, the framework achieves higher spatial precision in its selections. For instance, the USB cable and toothpaste tube were grasped with high success rates and showed low orientation errors, demonstrating ORACLE-Grasp’s effectiveness on both flexible and deformable items. The rubber duck, despite its irregular shape, also achieved high success rate with minimal position error. In contrast, the briefcase resulted in the lowest success rate and a relatively high orientation error. This may be attributed to its large overall size compared to the small handle, potentially requiring additional GRP iterations to identify a more focused grasp region. Similarly, the cardboard box showed the highest position error, likely due to its many potential graspable regions, which increase ambiguity in grasp selection. Despite such challenges, ORACLE-Grasp maintains high accuracy and reliability overall, with 88\% average success across all grasping attempts. Finally, while the Early Stopping mechanism does not influence the final accuracy, it serves as a critical efficiency module. In our experiments, it achieved a mean reduction in inference time of 27.48\% (±24.70\%), allowing the system to converge on an optimal grasp candidate more rapidly.

The impact of the depth-based grasp refinement module is quantified in the results. By comparing the positional RMSE with and without this refinement, we observe a significant enhancement in accuracy across diverse geometries. For instance, for the 'Coffee bag' and 'Cardboard box,' the RMSE was reduced by over 70\%. It is important to note that since the grasp refinement module operates on local depth data to refine the 3D contact point, it does not alter the grasp orientation $\theta$ derived during the reasoning phase. Furthermore, the reduction in positional error directly correlates with the high success rates observed; without grasp refinement, the initial 2D predictions occasionally target hollow or concave regions, which would result in physical grasp failure. These results validate grasp refinement as a critical bridge between high-level LMM reasoning and precise physical execution.

\section{Conclusions}

This work presented ORACLE-Grasp, a hybrid grasping framework that integrates the semantic reasoning capabilities of LMMs with classical computer vision techniques to achieve robust, task-relevant grasp prediction on unknown objects. By introducing a structured dual-prompt interaction, using SCP for high-level reasoning and GRP for spatial selection, we transform the inherently open-ended grasping task into a multiple-choice problem that LMMs can solve reliably. Through pre-cutting the image into candidate regions, aggregating results with IoU-based weighting and PCA, and applying depth-based clearance checks, ORACLE-Grasp effectively overcomes the spatial resolution limitations typical of LMMs. Additional mechanisms such as orientation refinement and early stopping contribute to both robustness and computational efficiency, enabling high success rates in both synthetic and real-world grasping experiments.

While traditional state-of-the-art grasp synthesis methods demonstrate high precision on known object distributions, they differ fundamentally from ORACLE-Grasp in both objective and infrastructure. Supervised models are typically 'expert students' of specific datasets; in contrast, ORACLE-Grasp acts as a 'reasoning generalist' that requires no prior training on specific object classes. Furthermore, standard geometric methods often prioritize mechanical stability over functional suitability. For instance, a purely geometric approach might successfully grasp a drill by its bit or a screwdriver by its shaft, which precludes subsequent task execution. ORACLE-Grasp prioritizes affordance alignment, using LMM-driven reasoning to identify handles and ergonomic interfaces. Finally, whereas many SOTA frameworks rely on dense 3D volumetric reconstruction or complex point cloud processing, our framework performs the core localization and orientation reasoning in the RGB domain. Depth information is utilized strictly for metric execution and local clearance checks. This shift from 3D-heavy modeling to LMM-driven visual reasoning allows ORACLE-Grasp to operate with lower computational overhead and greater flexibility in unstructured, zero-shot environments.

Despite these promising results, several limitations remain. The inference cost associated with LMMs is still substantial, posing challenges for real-time, high-frequency control loops. Future improvements could involve model compression, distillation, or leveraging more efficient architectures. Moreover, ORACLE-Grasp currently operates from a single camera viewpoint at inference time. While the model can semantically reason about objects and select plausible grasp regions, it does not actively seek alternative perspectives when critical features are occluded. Future extensions incorporating multi-view fusion or active exploration strategies could significantly enhance the system's adaptability, bringing robotic grasping closer to human-like flexibility.

While the current validation of ORACLE-Grasp focuses on single-object scenes to isolate its zero-shot affordance reasoning, the framework’s modular design is inherently extensible in future work to cluttered, multi-object environments. By parameterizing the SCP to accept specific target labels or complex human instructions, ranging from explicit commands like "pick-up the blue wrench" to implicit, task-oriented goals like "I need to tighten this bolt", the system can transition from autonomous principal object identification to targeted semantic search. In these scenarios, the LMM acts as a high-level planner that identifies the appropriate tool and its functional parts, allowing the subsequent GRP to ignore surrounding clutter and focus on local affordances. This evolution effectively bridges the gap between abstract natural language intent and low-level physical execution, paving the way for intuitive service robots capable of reasoning through the partial occlusions and complex spatial relationships of real-world settings.


\backmatter
\bmhead{Supplementary information}

Videos demonstrating the robot experiments, both in simulation and real-world environments, are available in the supplementary material.

\section*{Declarations}

\bmhead{Funding}

This work was supported by the Israel Science Foundation (No. 451/24).


\bmhead{Conflict of interest} The authors declare that there is no conflict of
interest or Competing interests.

\bmhead{Data availability} The test data collected in this work will be made available on request.

\bmhead{Code availability} Code will be made available on request.

\bibliography{ref}

\end{document}